\documentclass{article}

\usepackage{PRIMEarxiv}

\usepackage[utf8]{inputenc} % allow utf-8 input
\usepackage[T1]{fontenc}    % use 8-bit T1 fonts
\usepackage{hyperref}       % hyperlinks
\usepackage{url}            % simple URL typesetting
\usepackage{booktabs}       % professional-quality tables
\usepackage{amsfonts}       % blackboard math symbols
\usepackage{nicefrac}       % compact symbols for 1/2, etc.
\usepackage{microtype}      % microtypography
\usepackage{lipsum}
\usepackage{fancyhdr}       % header
\usepackage{graphicx}       % graphics
\graphicspath{{media/}}     % organize your images and other figures under media/ folder

\usepackage{xcolor, amsthm}
\usepackage{bbold}
\usepackage{algorithm}
\usepackage{algorithmic}
\usepackage{titletoc}
\usepackage{lettrine}
\usepackage{enumitem}
\usepackage{kpfonts}
\usepackage{anyfontsize}
\usepackage{subcaption}
\usepackage{comment}

\def \be*{\begin{eqnarray*}}
\def \e*{\end{eqnarray*}}
\def \beg{\begin{eqnarray}}
\def \en{\end{eqnarray}}

\def \bit{\begin{itemize}}
\def \eit{\end{itemize}}

\def \w {\widehat}

\theoremstyle{definition}

\title{Disentangled Deep Smoothed Bootstrap for \\ Fair Imbalanced Regression}
\author{
  Samuel Stocksieker \\
  Aix Marseille Université\\
  I2M, CNRS, Centrale Marseille\\ 
  Marseille, 
  France \\
   \And
  Denys Pommeret \\
  Aix Marseille Université\\
  I2M, CNRS, Centrale Marseille\\ 
  Marseille, 
  France \\
  \And
  Arthur Charpentier \\
  UQAM\\
  Montréal, Canada
}

\begin{document}
\maketitle

\begin{abstract}
Imbalanced distribution learning is a common and significant challenge in predictive modeling, often reducing the performance of standard algorithms. Although various approaches address this issue, most are tailored to classification problems, with a limited focus on regression. This paper introduces a novel method to improve learning on tabular data within the Imbalanced Regression (IR) framework, which is a critical problem. We propose using Variational Autoencoders (VAEs) to model and define a latent representation of data distributions. However, VAEs can be inefficient with imbalanced data like other standard approaches. To address this, we develop an innovative data generation method that combines a disentangled VAE with a Smoothed Bootstrap applied in the latent space. We evaluate the efficiency of this method through numerical comparisons with competitors on benchmark datasets for IR.
\end{abstract}

% keywords can be removed

\section{Introduction}

In various fields such as finance, economics, medicine, and engineering, regression techniques are frequently employed to establish relationships between features and a continuous target variable. However, real-world datasets often exhibit a significant imbalance in the distribution of the target variable, which poses specific challenges for traditional modeling methods \cite{krawczyk2016learning, fernandez2018learning}. This notion refers to certain ranges of values appearing more frequently than others, leading to an imbalance in learning. Typically, standard approaches treat all values as equally important and focus on minimizing the average error. As a result, rare and/or extreme values can have minimal influence on error reduction during the training phase due to their scarcity in the training set. This kind of modeling is unfair and biased: rare values are neglected in comparison with the majority values. Nevertheless, such values are often highly relevant for studies. 
The problem of learning from imbalanced data has been extensively studied in the context of classification (e.g. \cite{haixiang2017learning, fernandez2018smote}). However, the issue of Imbalanced Regression (IR) has received less attention, despite its significance to many real-world applications \cite{krawczyk2016learning}. Indeed, regression introduces additional challenges compared to classification: i) identifying rare/minority values is difficult due to the continuous nature of the distribution; ii) determining the degree of imbalance is not straightforward; iii) measuring the extent of rebalancing is also difficult; and iv) generating synthetic data requires the creation of a new value for the target variable, associated with the new synthetic features.

Three categories of non-exclusive solutions address this issue: pre-processing, in-processing, and post-processing techniques \cite{branco2016survey}. Pre-processing techniques aim to balance the target variable's distribution before model training, while in-processing techniques modify the learning process itself to better manage data imbalances. %Post-processing techniques adjust model predictions after they have been generated to enhance overall performance.  
Preprocessing solutions are often the preferred choice for tabular data due to their universality, allowing the use of any standard regression technique after resampling \cite{he2013imbalanced}. 
Preprocessing techniques primarily rely on resampling by generating synthetic data. Various methods have been suggested to create synthetic observations from the original data space, such as the widely recognized SMOTE algorithm \cite{Chawla2002Smote}. Another approach involves using a method that embeds observations into a latent space from which data can be generated, like Variational Autoencoders (VAEs). VAEs are a data generation method that employs a neural network to encode observations into a continuous latent space. The probabilistic structure of this latent space can then be utilized to simulate new data. 
%This approach has proven to be quite effective in generating unstructured data, especially due to its ability to capture nonlinear relationships. Our objective here is to generate artificial rare values while taking into account the observed correlations and, most importantly, the dependency between the target variable and the features.  Indeed, when using a synthetic data generator, regression requires generating new values for the continuous target variable, unlike in classification.
%In addition, it also enables us to handle mixed data, which is a challenge in data generation.
The key idea of this paper is to propose a new data representation space to enable more relevant synthetic data generation, especially by accounting for categorical data and nonlinear correlations. In particular, we propose a suitable framework for the application of the Smoothed Bootstrap, which is generally not the case in the initial data space.
%Indeed, with this kind of data generation, there is no guarantee of performance if the conditions for its application (continuous distribution) are not fulfilled.
%Finally, the purpose of combining a Smoothed Bootstrap with a VAE is
We aim to combine a Smoothed Bootstrap with a VAE to address the potential lack of robustness of the VAE in inferring rare values, as underlined by \cite{wang2024variational}. Furthermore, generations based on VAE are efficient as long as the latent distributions are well-adapted. 
In this paper, we propose a simple and effective adaptation of disentangled $\beta-$VAEs to tailor their training and generation specifically for the problem of IR on structured data. 
Our main contributions can be outlined as follows: 
\bit
\item Introducing a novel loss function for training Disentangled VAEs in the context of IR 
\item Using a free-parameter weighting scheme to manage the sampling of rare values when the target variable is continuous (regression) 
\item Presenting a new synthetic data generator that combines the learning power of disentangled VAEs with a non-parametric generator: the (Disentangled) Deep Smoothed Bootstrap. 
\eit
To our knowledge, disentangled VAEs have never been used for synthetic data generation applied to tabular datasets. This application highlights a new use case for disentangled VAEs. Moreover, unlike the imbalanced regression techniques in the benchmark, our approach can handle mixed data types and account for nonlinear correlations, reproducing them during generation. This is made possible by the transformation into a latent representation space.
The paper is organized as follows. 
Section \ref{RelatedWorks} presents various state-of-the-art works, highlighting their shortcomings and how they differ from our proposal.
Section \ref{proposition} introduces our DSB (for Deep Smoothed Bootstrap) approach, which combines a disentangled VAE adapted to IR with an alternative method generation: the Smoothed Bootstrap. A scheme of our approach is presented in Figure \ref{DisDSB_Scheme}.
Section \ref{Experiments} presents real-world applications. 
Finally, Section \ref{Discussion} discusses the proposed method and presents some perspectives.

\section{Related Work}
\label{RelatedWorks}
\paragraph*{Tabular Imbalanced Regression} 
The initial efforts on IR were based on a utility function that enabled the binarization of the problem. Indeed, \cite{ribeiro2011utility} initially suggested using a relevance function to assign a value to each instance of the target variable $Y$. Subsequently, the majority and minority values are identified using a user-defined threshold. This initial approach facilitates the adaptation of existing solutions within such as the well-known SMOTE algorithm to regression \cite{torgo2013smote},  Gaussian Noise \cite{branco2016ubl, branco2017smogn, song2022distsmogn} or SMOTE extensions \cite{moniz2018smoteboost, camacho2022geometric}. 
While this approach has paved the way for the first solutions to IR, it has the drawback of losing information about the target variable. Moreover, these utility-theoretic approaches are highly sensitive to the arbitrarily chosen utility function and associated threshold. A more recent work suggested an alternative approach without binarizing the problem: GOLIATH algorithm \cite{stocksieker2023goliath}.
Lastly, as for classification, standard performance metrics are also biased and need to be reevaluated in the case of an imbalanced distribution \cite{branco2016survey, ribeiro2020imbalanced}. 
Generating data from the initial space involves two major challenges: 1) the need to identify existing correlations in order to respect and reproduce them, and 2) the handling of categorical data.
Finally, existing methods seek to generate synthetic data in the data space, with varying degrees of complexity. However, these techniques are subject to two important conditions: i) being able to identify potentially complex dependencies in the data, and ii) being able to reproduce them during generation.

\paragraph*{Deep Imbalanced Regression}
More recently, the concept of IR has been extended to image and text (NLP) data, such as estimating the age of individuals from photos, leading to the development of Deep Imbalanced Regression (DIR) \cite{yang2021delving}. Given the effectiveness of deep learning approaches on this type of data, it is natural that new methodologies have emerged. For instance, \cite{yang2021delving} proposed using kernel density estimations to smooth the distributions of variables, thereby enhancing learning with a neural network when the target variable is continuous and imbalanced. 
Building on this idea of smoothing via kernel density estimation, \cite{ding2022deep} introduced the techniques of "CORrelation ALignment" and "class-balanced re-weighting" to enhance results. Furthermore, \cite{gong2022ranksim} proposed a method to improve the performance of deep regression models in IR  scenarios by incorporating a ranking similarity regularization into the loss function. On the other hand, \cite{keramati2023conr} suggested a technique known as a \textit{contrastive regularizer}, which is derived from Contrastive Learning. \cite{sen2023dealing} addressed the issue by employing a logarithmic transformation combined with an artificial neural network. 
Building on previous work in classification  \cite{ai2023generative, utyamishev2019progressive}, \cite{wang2024variational} proposed a new approach using Variational Autoencoders (VAEs) to address DIR problems. This method modifies traditional VAEs, which usually make independent inferences for each observation (assuming i.i.d. latent representations), by incorporating similarities between observations.

\section{A Deep Smoothed Bootstrap for Imbalanced Regression}\label{proposition}

Let $\boldsymbol{x}=(x_{ij})_{i=1,\ldots,n ; j=1,\ldots,p} \in \mathcal{X} \subset \mathbb{R}^{n \times p}$ be a dataset composed of $n$ observations and $p$ variables where $x_{ij}$ represents the $j$-th variable for the $i$-th observation. Let $y=(y_{i})_{i=1,\ldots,n} \in \mathcal{Y} \subset \mathbb{R}^n$ denote the associated target variable, which is continuous and we assume to be imbalanced.
We propose to construct the Deep Smoothed Bootstrap in three steps. The first one is to use a ($\beta$-)VAE specific to the regression context, i.e., in a supervised framework. The second one is to adapt this approach to IR by modifying the loss function (in-processing). This adaptation includes adjusting the latent space, with a disentanglement, to facilitate the application of the Smoothed Bootstrap. Finally, the last step involves redefining the mode of synthetic data generation of the VAE by applying the Smoothed Bootstrap in the latent space, which provides an ideal framework for its application.

\paragraph{$\beta$-VAE for Regression}

Classical autoencoders are a type of neural network designed to learn efficient representations of data, typically for purposes such as dimensionality reduction or feature extraction. They consist of two main components: an encoder and a decoder. The encoder compresses the input data into a lower-dimensional representation: a "latent space." The decoder then attempts to reconstruct the original data from this compact representation. The objective of training an autoencoder is to minimize the difference between the input data and its reconstruction, thus ensuring that the learned latent space effectively captures the most important features of the data. 
Autoencoders are useful in various applications, such as noise reduction, anomaly detection, and data compression. However, traditional autoencoders often struggle to learn meaningful or interpretable representations. They can overfit the training data and do not explicitly enforce any particular structure on the latent space. Moreover, traditional autoencoders do not support data generation, particularly due to the lack of regularity in the latent space.
To address these limitations, Variational Autoencoders (VAEs) were introduced. VAEs are a probabilistic extension of autoencoders that incorporate principles from Bayesian inference. Unlike traditional autoencoders, VAEs model the latent space as a probability distribution, typically Gaussian. The encoder maps input data to the parameters of the theoretical distributions (mean and variance for Gaussian ones), rather than to a single point in the latent space. The decoder then reconstructs the data from samples drawn from this distribution. The key innovation of VAEs is their ability to produce a smooth and continuous latent space, where similar data points are represented by nearby latent vectors. This property is particularly valuable for generative tasks, as it enables the generation of new, unseen data by sampling from the latent space.
% Regression is a method used to understand and predict the relationship between a target variable and other variables. It often relies on estimating a function $f$ such that $\mathbb{E}(y|X) = f(X)$. 
However, when using a traditional VAE in a regression context, the target variable $Y$ should not be considered as a feature $X$. As a result, we suggest a mixed VAE approach: this VAE takes both the input features $X$ and the target variable $Y$, but with a specific weighting for $Y$.
The global loss function is typically defined as follows:
\vspace{-0.5cm}
\begin{align*}
\mathcal{L}(\theta, \phi, \boldsymbol{x}, y)  = \beta_x \mathbb{E}_q[\log \, p_\theta(\boldsymbol{x}|\boldsymbol{z})] + \beta_y \mathbb{E}_q [\log \, p_\theta (y|\boldsymbol{z})]   - \beta_{KL} D_{KL}(q_\theta(z|\boldsymbol{x,y})\|p_\theta(\boldsymbol{z})) ,
\end{align*}
\vspace{-0.5cm}

where $\beta_x$ (resp. $\beta_y$) and $\beta_{KL}$ are the weights associated to % $\mathbb{E}_q [log \, p_\theta (\boldsymbol{x}|\boldsymbol{z})]$ (resp.  $\mathbb{E}_q [log \, p_\theta (y|\boldsymbol{z})]$) represents 
the reconstruction loss function for $\boldsymbol{x}$ (resp.  $y$) and %$D_{KL}(q(z|x), p(z|x))$ represents 
the Kullback-Leibler Divergence for regularization. Note that $\beta_{KL}$ = 1 is the special case of the classical VAE and a $\beta-$VAE, otherwise, it refers to the first proposed disentangled VAE \cite{higgins2017beta}.
Here $q_\theta(\boldsymbol{z}|\boldsymbol{x,y})$  represents the distribution of latent variables $\boldsymbol{z}$ given the input data $\boldsymbol{x,y}$ and $p_\theta(\boldsymbol{z})$ denotes the prior distribution, which is often chosen to be Gaussian. The advantage of using a VAE  is to ensure the regularity (continuity) of the latent space through penalization, thereby facilitating coherent data generation. %This would not be the case with a classic autoencoder. 
%Indeed, t
This is not the case with a traditional autoencoder, which lacks the ability to enforce the smoothness of the latent space. This smoothness is crucial for generating coherent data.

\paragraph{Why and how to adapt a $\beta$-VAE for Imbalanced Regression Modeling ?}
% \subsubsection{The limitations of standard ($\beta$-)VAEs}
It has been shown that the Mean Squared Error (MSE) loss function is not effective for handling IR  \cite{ren2022balanced}. This is because it inherently favors frequent values, which significantly reduce the loss function, and as a result, neglects rare values. \cite{stocksieker2024boarding} also showed that standard ($\beta$-)VAEs, using a multivariate MSE, tended to neglect rare values.
As suggested by \cite{wang2024variational}, standard ($\beta$-)VAEs need to be adapted to the problem of IR to be effective.
By conducting a thorough analysis of rare value modeling with a $\beta$-VAE, we observed the following insights:
\bit
\setlength\itemsep{0.1em}
\item The variance ($\sigma$ parameter of the $\beta$-VAE) of the latent Gaussians increases for rare values. 
%The Spearman coefficient between the variance of the latent distributions ($\sigma$) and the rarity level, measured by the inverse of the distribution density ($w_i$), is on average 54\%: the lower the observation, the higher its variance.
\item The reconstruction error increases with the rarity of the observation
%, measured by the inverse of the distribution density ($w_i$). The Spearman coefficient between the reconstruction RMSE and $w_i$ is 13\%. This is naturally related to the increase in the variance of the latent distributions.
\item The prediction error increases with the rarity of the observation
%, measured by the inverse of the distribution density ($w_i$). The Spearman coefficient between the reconstruction RMSE and $w_i$ is 24\%. This observation was also presented in \cite{yang2021delving}: "\textit{the error distribution correlates with label density distribution}".
\eit

As observed by \cite{wang2024variational}, the standard ($\beta$-)VAE is not suited to handle IR. Since the multivariate MSE gives equal weight to all observations, rare values are mechanically neglected because they do not significantly reduce the value, unlike frequent values. Therefore, it is important initially to give more weight to rare values during the training phase of the algorithm (as in \cite{yang2021delving}).

% \subsubsection{A Balanced loss function for Imbalanced Regression}
To address the previous issues, we propose using a balanced loss function that considers the frequency of the target variable $Y$. The aim is to train a model capable of generating new synthetic data for the rare values of $Y$. In classification tasks, rare values are directly identifiable as they belong to one or more minority classes. However, in regression, where the support of the target variable is continuous, defining rare values presents an initial challenge.
To address this problem, we propose weighting the observations during the training phase by the inverse of the empirical density of $Y$. In other words,  rarer values receive higher weights, while more frequent values receive lower weights. Additionally, we recommend adjusting these weights using a parameter $\alpha>0$ (to be discussed later). 
The global loss function thus becomes:

\vspace{-0.5cm}
\begin{align}
\label{Bal_loss}
    \mathcal{L}(\theta, \phi, \boldsymbol{x}, y) =  \beta_x \mathbb{E}_q[\log \, p_\theta(\boldsymbol{x}|\boldsymbol{z})]  +  \frac{\beta_y}{\w f(y)^\alpha} \mathbb{E}_q [\log \, p_\theta (y|\boldsymbol{z})]   - \beta_{KL} D_{KL}(q_\theta(\boldsymbol{z}|\boldsymbol{x,y})\|p_\theta(\boldsymbol{z})),
\end{align}
\vspace{-0.5cm}

%Since the density of Y is unknown, we suggest estimating it using a 
\noindent where $\w f_y$ is a kernel density estimator:$\displaystyle \widehat{f}(y) = \frac{1}{n h} \sum_{i=1}^{n} K\left(\frac{y - y_i}{h}\right), $
with \( K \) the kernel function and \( h \) the bandwidth. This form of weighting is close to the DIR \cite{yang2021delving} technique proposing to weight by the inverse of the kernel estimator as well.

\paragraph{A Smoothed Bootstrap for Data Generation}

% \subsubsection{Data Generation Process}
% As previously mentioned, the main advantage of Variational Autoencoders (VAEs) is their ability to generate data that closely resembles the training data. Typically, a standard VAE -using Gaussian distributions- focus on calibrating the parameters $\mu_i$ and $\sigma_i$ of a latent normal distribution to align with each observation $x_i$. 
% Calibrating a ($\beta$-)VAE in Imbalanced Regression is challenging because algorithms often struggle to model rare values. This difficulty is also encountered with classical autoencoders. 
In the previous section, we proposed an initial adjustment, a balanced MSE, to improve the ($\beta$-)VAE's learning process. Despite this, the variance $\sigma_i$ assigned to rare values might still be unstable, due to the lack of observations, which means the generated data may not accurately reflect the original data. 
%This occurs because 
The ($\beta$-)VAE creates a latent representation for each observation, calibrating the parameters of each latent variable independently. These latent representations are treated as identically and independently distributed (i.i.d.) \cite{wang2024variational}.
It is important to emphasize that the data generation process focuses specifically on rare values. The objective is to create a training sample that includes both "majority" values that are observed and rare values that are either observed or generated.

Here, we suggest a second level of adjustment to enhance the generation of rare data. The first step involves defining the drawing weights, which are used for sampling. These weights are also used to rebalance the learning process and enable better modeling of rare values. Inspired by \cite{stocksieker2023goliath}, we suggest to define these weights as follows: $ \displaystyle \omega_i:= \frac{1}{\widehat{f}_Y(y_i)^\alpha}$.
We then suggest an alternative approach for generating data with the VAE. Instead of using the conventional method (from latent Gaussian distributions), we propose applying a Smoothed Bootstrap \cite{silverman1987bootstrap, hall1989smoothing, de1992smoothing}. This method is applied to the $n\times q$ matrix of values $\mu$, which represents the mean value of the new representation of $(X,y)$ in the latent space of dimensionality $q$. %This approach enables a more nuanced generation of data, which is particularly beneficial for handling rare values.
The Smoothed Bootstrap (SB) method involves generating samples from kernel density estimators of the distribution. This process can be decomposed into two steps:
\begin{enumerate}
\setlength\itemsep{0.1em}
    \item Seed Selection, using the weights $\omega_i$; 
    \item 
    Random variables are generated, using  kernel density estimators.
    %Random Noise Addition, from the kernel density estimator. 
    %is added to the seed to create a new sample. This step involves generating a kernel on $\mu$.
\end{enumerate}
Building on this, we propose to use the following mixture of multivariate kernels (e.g. Gaussian) to generate synthetic data:  $\displaystyle  g_{Z^*}(z^*| \mathbf{\mu}) = \displaystyle \sum_{i=1,\ldots,n} \omega_i K_i(z^*,\mathbf{\mu})$, 
where $(K_i)_{i =1,\ldots,n}$ is a collection of kernel, and  $(\omega_i)_{i=1,\ldots,n}$ is a sequence of positive weights with
$\sum_{i=1,\ldots,n} \omega_i=1$. 
The SB consists in generating $z^*$ from $g_{Z^*}$, the index $^*$ denoting  synthetic data. In practice, we have used Gaussian kernels to which a variance matrix must be associated. The Smoothed Bootstrap has the advantage of not requiring additional parameters. 
Unlike($\beta$-)VAE, this method has the following advantages: i) using the neighborhood of the observation in the generation;  ii) allowing control of the generation level (with a hyperparameter $hmult$ of level noise) and iii) A default smoothing parameter already optimized \cite{Scott92, silverman1986density} that ensures consistency in the estimation process \cite{de2008multivariate}. Our method focuses on the first case (tabular data) and allows for the generation of synthetic data to adjust the sample. This preprocessing solution enables the use of any prediction algorithm afterward. Such a solution is not recommended for unstructured data, which requires significant computational time for training.

\paragraph{The advantage of a disentangled representation}
With standard usage, the construction of our Smoothed Bootstrap only requires a parameter: the level of noise. Another significant parameter is the smoothing parameter, and we utilize already optimized smoothing matrices that demonstrate convergence properties under specific regularity and probability conditions. This approach avoids the arbitrary choice of the bandwidth window, which is arbitrarily fixed in \cite{yang2021delving}.
However, the suggested smoothing estimator (in \cite{silverman1986density} or \cite{Scott92}) is based on the empirical variance-covariance matrix, which may be inadequate if the latent variables exhibit non-linear correlations. Nothing guarantees that the generation takes into account the non-linear links that may exist in the latent space.

To address this issue, we propose adjusting the latent space to ensure an ideal framework for the application of the Smoothed Bootstrap. More specifically, we aim to obtain a latent space where the variables are not correlated with each other. Seeking to construct a latent space with independent dimensions is precisely the aim of disentangled VAEs. However, according to \cite{locatello2019challenging}, disentangled VAEs do not guarantee the non-correlation of latent representation. According to this comparative analysis, the best approach to achieve a non-correlated space (in terms of total correlation) is penalization by covariance using DIP-VAE \cite{kumar2017variational}. However, since covariance is unitless, we prefer to use the correlation coefficient to avoid scaling effects between different variables. 
We therefore propose to minimize the correlation between latent variables within the loss function by adding a penalty. Let $r$ be the standard correlation coefficient defined as follows: $ \displaystyle r(z_i,z_j) = \frac{Cov(z_i,z_j)}{\sigma_{z_i} \sigma_{z_j}}$, our final loss function thus becomes, extended the previous form \ref{Bal_loss}:
\vspace{-0.5cm}
% \small{
\begin{align}
\label{Final_loss}
    \mathcal{L}(\theta, \phi, \boldsymbol{x}, y)  =  \beta_x \mathbb{E}_q[\log \, p_\theta(\boldsymbol{x}|\boldsymbol{z})]  +  \frac{\beta_y}{\w f(y)^\alpha} \mathbb{E}_q [\log \, p_\theta (y|\boldsymbol{z})]   - \beta_{KL} D_{KL}(q_\theta(\boldsymbol{z}|\boldsymbol{x,y})\|p_\theta(\boldsymbol{z}))   + \beta_{corr}  \underset{\substack{i,j \\ i \ne j}}{\sum} r^2 (z_i, z_j)
\end{align}
% }
\vspace{-0.5cm}

\section{Experiments}\label{Experiments}

% \subsection{Data description and experimental setup}
To evaluate our method on real-world datasets, we compare our results with benchmark datasets for IR sourced from the study by \cite{branco2019pre}\footnote{The dedicated repository "Data Sets for Imbalanced Regression Learning" is available at this address: \url{https://paobranco.github.io/DataSets-IR/} }. Two datasets are numerical (\textit{cpuSm} and \textit{boston}) and two datasets are mixed (\textit{pricingGame}) and \textit{availPwr}) i.e., with categorical covariates that are transformed with a one-hot encoding.
The datasets are:
\bit[noitemsep, topsep=0pt]
\setlength\itemsep{0.1em}
\item \textit{cpuSm}: 8192 samples and 16 variables (3 float, 11 integer, and 6 categorical).
\item \textit{pricingGame}: 100000 samples and 16 variables (1 float, 9 integer, and 6 categorical).
\item \textit{availPwr}: 1711 samples and 15 variables (2 float, 6 integer, and 7 categorical).
\item \textit{boston}: 506 samples and 14 variables (11 float and 3 integer).
\eit
% \subsection{Experimental setup}
For each dataset, we construct a training set and a test set using uniform random sampling. 
We then build four models: a standard autoencoder, a standard $\beta$-VAE, a weighted $\beta$-VAE (based on \ref{Bal_loss}), and our disentangled weighted $\beta$-VAE (based on \ref{Final_loss}). 
Once the models are built, we generate several synthetic samples from the same seed drawing with the weights defined previously, $\omega_i$. 
% \subsection{Ablation study and Benchmark}
Several sets of training data are constructed, including ablation components and state-of-the-art solutions:
\begin{itemize}[noitemsep, topsep=0pt]
\item \textit{Baseline}: The initial training set.
\item \textit{OS}: Oversampling applied on the training set, using the drawing weights referenced in $\omega_i$.
\item \textit{SB+AE}: A combination of Smoothed Bootstrap with an autoencoder. This method should be avoided because the autoencoder does not ensure the regularity of the latent space, which theoretically hinders the application of the Smoothed Bootstrap.
\item \textit{BVAE}: Natural generation with the standard $\beta$-VAE.
\item \textit{kBVAE}: A combination of Smoothed Bootstrap with a standard $\beta$-VAE.
\item \textit{BVAEw}: Natural generation of a weighted $\beta$-VAE trained with the loss function defined in \ref{Bal_loss}.
\item \textit{kBVAEw}: A combination of Smoothed Bootstrap with a weighted $\beta$-VAE trained with the loss function defined in \ref{Bal_loss}. 
\item \textit{DSB} is the Deep Smoothed Bootstrap: with the correlation maximization penalty added in the loss function \ref{Final_loss}.
\item \textit{TVAE}: A Tabular Variational Autoencoder \cite{ctgan}, trained with the same number of epochs as our models, using \textit{Synthetic Data Vault} (SDV)  Python package \cite{SDV}.
\item \textit{CTGAN}: A conditional Tabular Generative Adversarial Network, also trained with the same number of epochs as our models, using the SDV Python package.
\item \textit{CopGAN}: A Copula-Generative Adversarial Network, trained with the same number of epochs as our models, using the SDV Python package.
\item \textit{ILRro}: Oversampling applied to the training set, using the UBL approach \cite{branco2016ubl} from the \textit{imbalancedlearningregression} Python package \cite{wu2022imbalancedlearningregression}.
\item \textit{ILRsmote}: The SMOTE for Regression method \cite{torgo2013smote}, applied to the training set, using the UBL approach from the \textit{imbalancedlearningregression} Python package.
\item \textit{ILRgn}: The Gaussian Noise for Regression method \cite{branco2017smogn}, applied to the training set, using the UBL approach from the \textit{imbalancedlearningregression} Python package.
\end{itemize}
The idea of generating from a latent space can be found in \cite{tang2008generation} and \cite{martinez2012sneom}. However, the authors propose to construct a latent (factorial) space with a Principal Components Analysis, which does not capture non-linear correlations and therefore does not solve the problem. A Smoothed Bootstrap combined with principal components and kernel principal components were tested, but the results were removed due to poor performance.\\
% \subsection{Protocol}
The next step involves using the various training datasets to predict the target variable $Y$ of the test set, with the initial training dataset serving as the baseline. To ensure robust results, we perform the comparison on 10 train-test samples, following a K-fold approach. It is important to note that the training sets are mixed, meaning they are composed of a blend of original data (primarily for frequent observations) and synthetic data (primarily for rare values). Similarly, to prevent our results from being dependent on specific learning algorithms, we utilize 10 models from the \textit{autoML of the H2O package} \cite{H2OAutoML20}. These models employ a variety of algorithms, including Distributed Random Forest, Extremely Randomized Trees, Generalized Linear Model with regularization, Gradient Boosting Model, Extreme Gradient Boosting, and a fully connected multi-layer artificial neural network. In our experiments, our hyperparameters are set as follows: $\beta_x=1, \beta_y=3, \alpha = 1, \beta_{KL}=1e-5 \text{ and } \beta_{corr}=1$. A sensitivity analysis to the $\beta_{corr}$ parameter was carried out and showed empirically that the value of 1 gave better results on simulations.

% \subsection{Results}
\begin{figure}[ht]
\centering
\begin{minipage}[t]{0.6\textwidth}
\centering
\small{
\begin{tabular}{|p{1.6cm}||p{1.6cm}|p{1.6cm}|p{1.6cm}|p{1.6cm}|}
 \hline
 Training \ Dataset	&cpuSm	&pricingGame	&availPwr	&Boston\\
 \hline
Baseline	&2.1 $\pm$ 0.03	&35.3 $\pm$ 0.35	&4.3 $\pm$ 0.12	&2.3 $\pm$ 0.11\\
kAE	        &2 $\pm$  0.04      &34.4 $\pm$ 0.27	&4.2 $\pm$ 0.17	&2.2 $\pm$ 0.16\\
OVAE	    &2.1 $\pm$ 0.04	    &35.6 $\pm$ 1.06	&4.4 $\pm$ 0.21	&2.2 $\pm$ 0.13\\
BVAE        &2.1 $\pm$ 0.02    &34.6 $\pm$ 0.67	&4.4 $\pm$ 0.15	&2.2 $\pm$ 0.13\\
kBVAE       &2.1 $\pm$ 0.03	    &34.3 $\pm$ 0.3	&4.2 $\pm$ 0.15	&2.2 $\pm$ 0.16\\
BVAEw       &2.1 $\pm$ 0.04	    &35.5 $\pm$ 1.19	&4.3 $\pm$ 0.19	&2.2 $\pm$ 0.15\\
kBVAEw      &2.1 $\pm$ 0.04	    &34.4 $\pm$ 0.32	&4.3 $\pm$ 0.13	&2.2 $\pm$ 0.15\\
\textbf{DSB} (ours)	&\textbf{2 $\pm$ 0.04}      &\textbf{33.9 $\pm$ 0.3}	&\textbf{4.1 $\pm$ 0.1}	&\textbf{2.1 $\pm$ 0.16}\\
OS          &2.1 $\pm$ 0.03	&34.9 $\pm$ 0.24	&4.3 $\pm$ 0.11	&2.3 $\pm$ 0.11\\
TVAE        &2.2 $\pm$ 0.1	&34.9 $\pm$ 0.26	&4.7 $\pm$ 0.26	&2.4 $\pm$ 0.14\\
CTGAN       &2.3 $\pm$ 0.09 &34.5 $\pm$ 0.46	&6.1 $\pm$ 2.96	&2.5 $\pm$ 0.16\\
CopGAN      &2.4 $\pm$ 0.07	&34.4 $\pm$ 0.49	&5.6 $\pm$ 0.35	&2.6 $\pm$ 0.19\\
ILRro       &2.1 $\pm$ 0.03	&35.7 $\pm$ 0.28	&4.3 $\pm$ 0.16	&2.3 $\pm$ 0.15\\
ILRsmote	&2.4 $\pm$ 0.1	&35 $\pm$ 	0.33    &5.1 $\pm$ 0.1	&2.3 $\pm$ 0.15\\
ILRgn       &2.1 $\pm$ 0.04	&34.9 $\pm$ 0.3	&4.4 $\pm$ 	0.1     &2.2 $\pm$ 0.13\\
 \hline
\end{tabular}
}
\captionof{table}{Experiments results}
\label{table_mean}
\end{minipage}
\hfill
\begin{minipage}[c]{0.39\textwidth}
\centering
\includegraphics[width=0.7\textwidth]{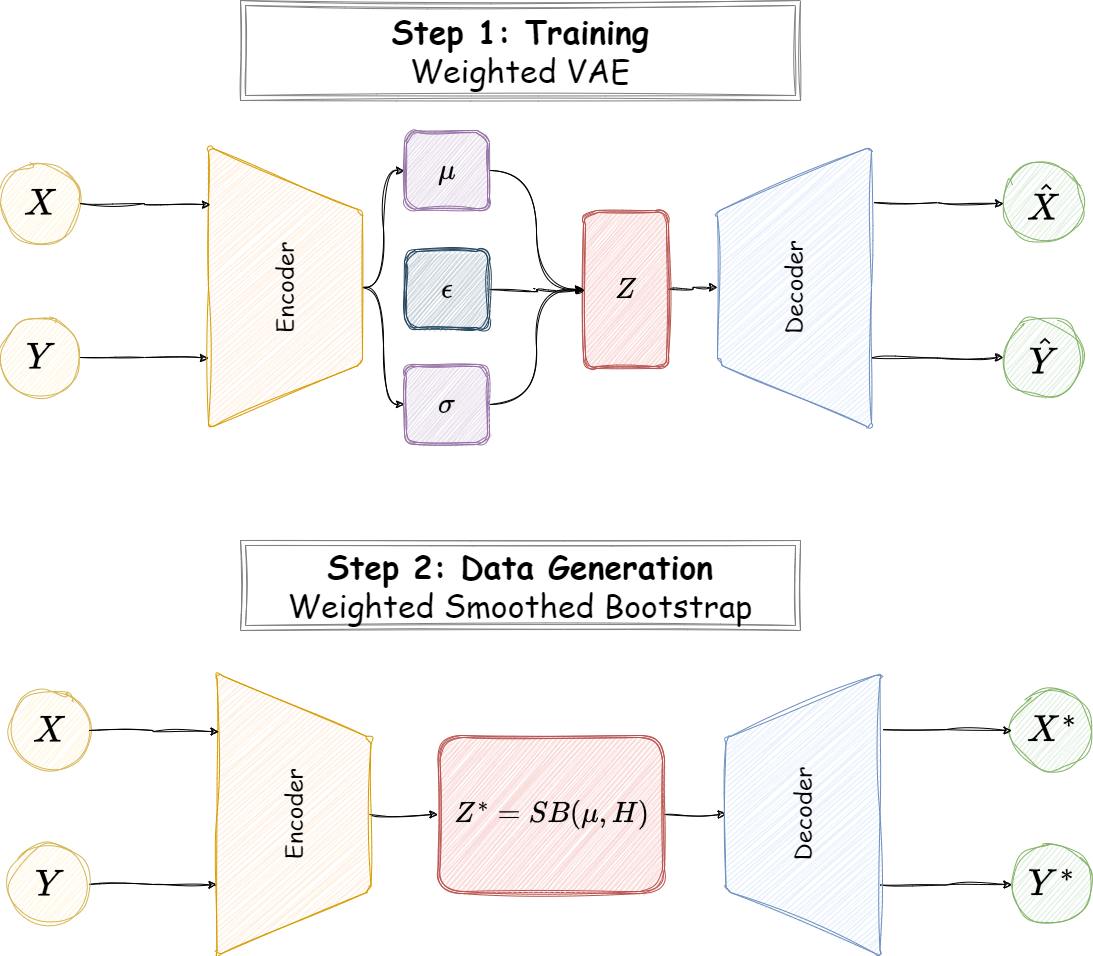}
\captionof{figure}{Scheme of our approach Disentangled DSB}
\label{DisDSB_Scheme}
\end{minipage}
\end{figure}

\begin{figure}[ht]
\centering
\begin{subfigure}{0.49\textwidth}
  \includegraphics[width=\textwidth]{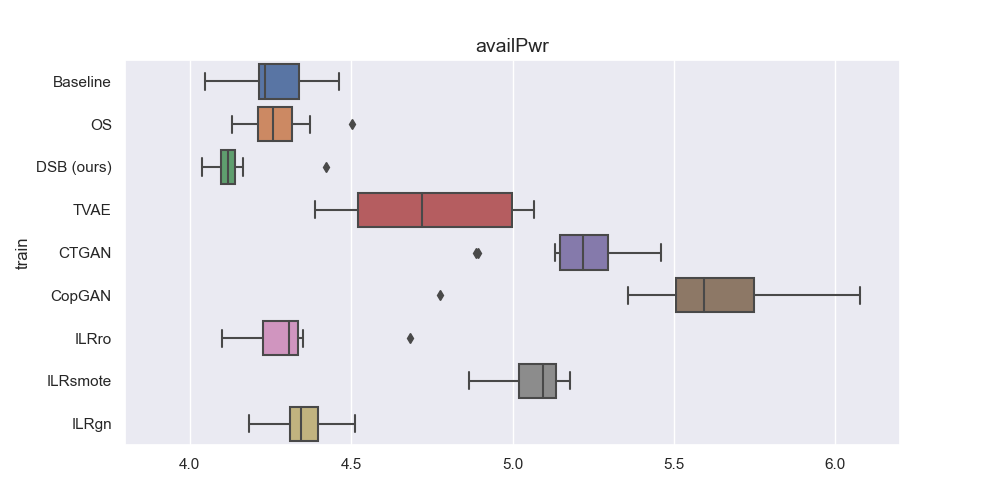}
  % \caption{Pairplot}
  % \label{IlluSB_pairplot}
\end{subfigure}
\begin{subfigure}{0.49\textwidth}
  \includegraphics[width=\textwidth]{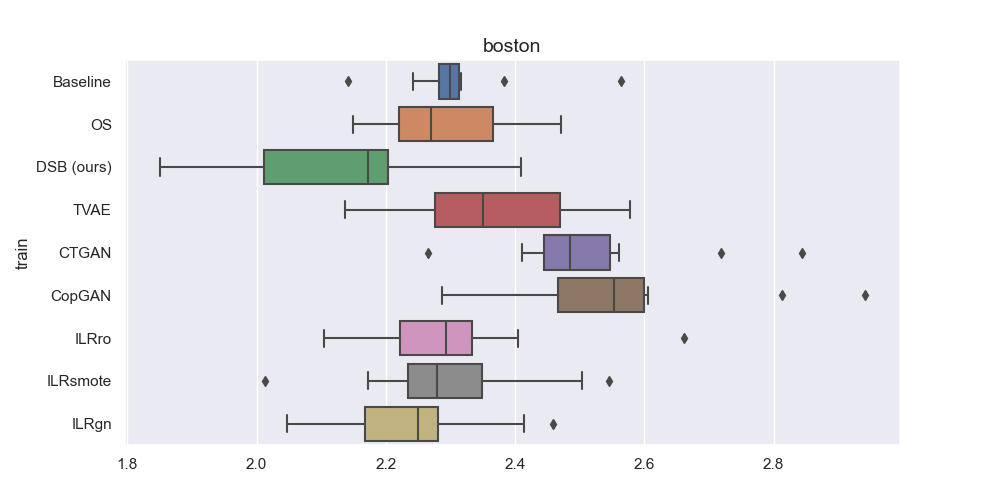}
  % \caption{Reference correlation matrix}
  % \label{RefMC}
\end{subfigure}
% \hfill
\begin{subfigure}{0.49\textwidth}
  \includegraphics[width=\textwidth]{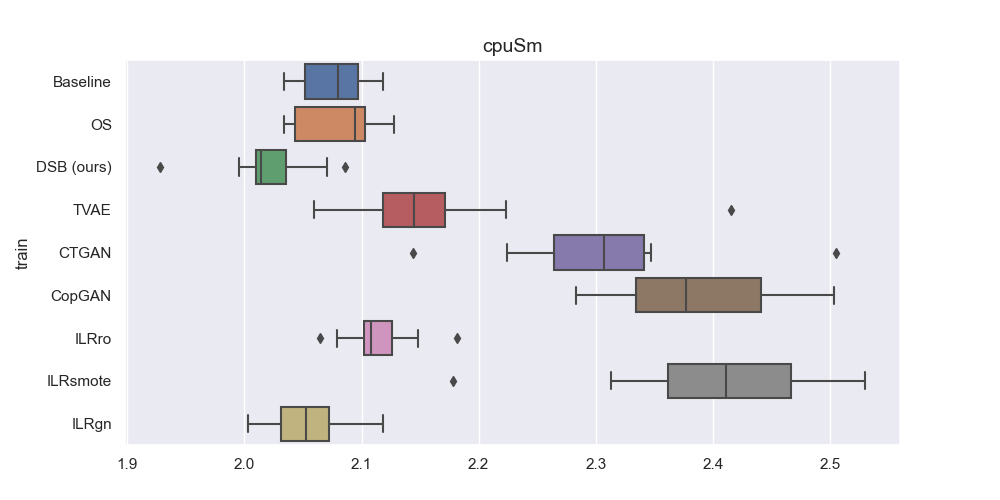}
  % \caption{Our \texttt{KurtHGR} correlation matrix}
  % \label{PearsonMC}
\end{subfigure}
\begin{subfigure}{0.49\textwidth}
  \includegraphics[width=\textwidth]{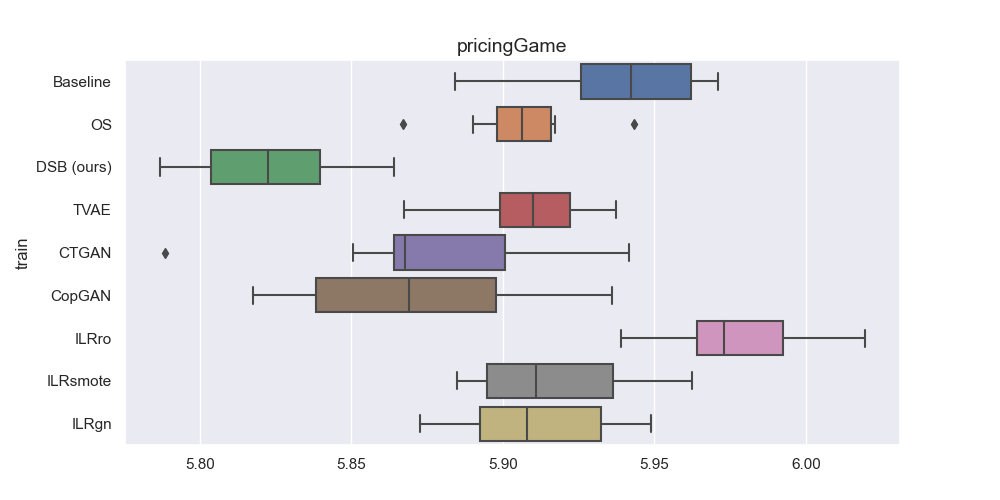}
  % \caption{Our \texttt{KurtHGR} correlation matrix}
  % \label{PearsonMC}
\end{subfigure}
\caption{Comparison of our DSB approach (in green) versus the state of the art}
\label{Benchmark_res}
\end{figure}

\begin{figure}[ht]
\centering
\begin{subfigure}{0.49\textwidth}
  \includegraphics[width=\textwidth]{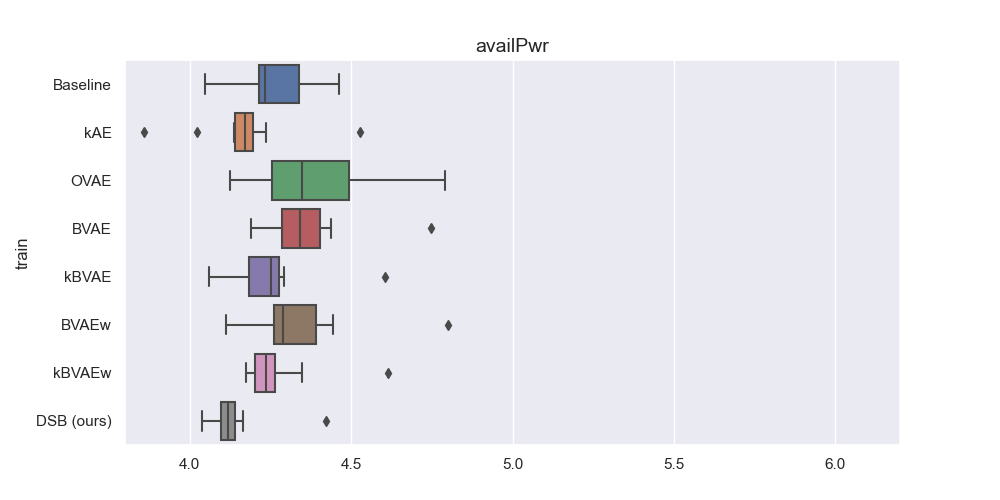}
  % \caption{Pairplot}
  % \label{IlluSB_pairplot}
\end{subfigure}
\begin{subfigure}{0.49\textwidth}
  \includegraphics[width=\textwidth]{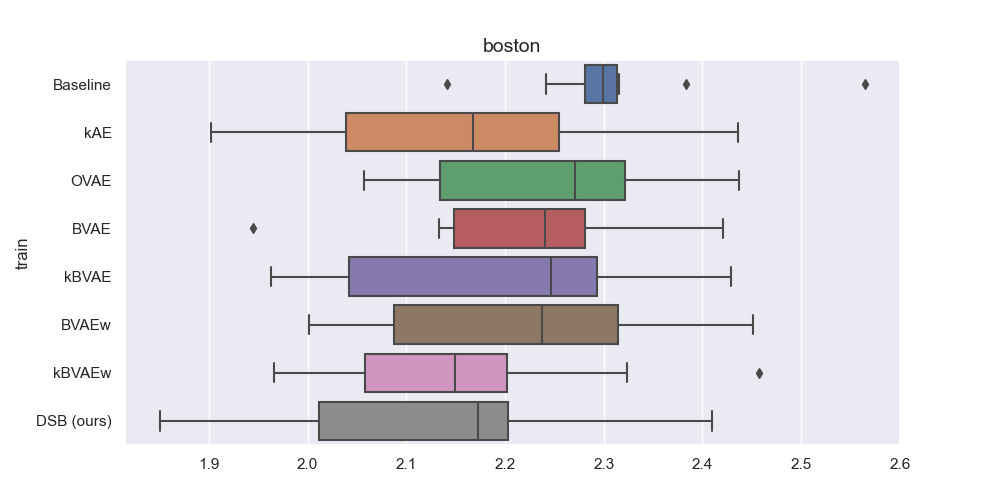}
  % \caption{Reference correlation matrix}
  % \label{RefMC}
\end{subfigure}
% \hfill
\begin{subfigure}{0.49\textwidth}
  \includegraphics[width=\textwidth]{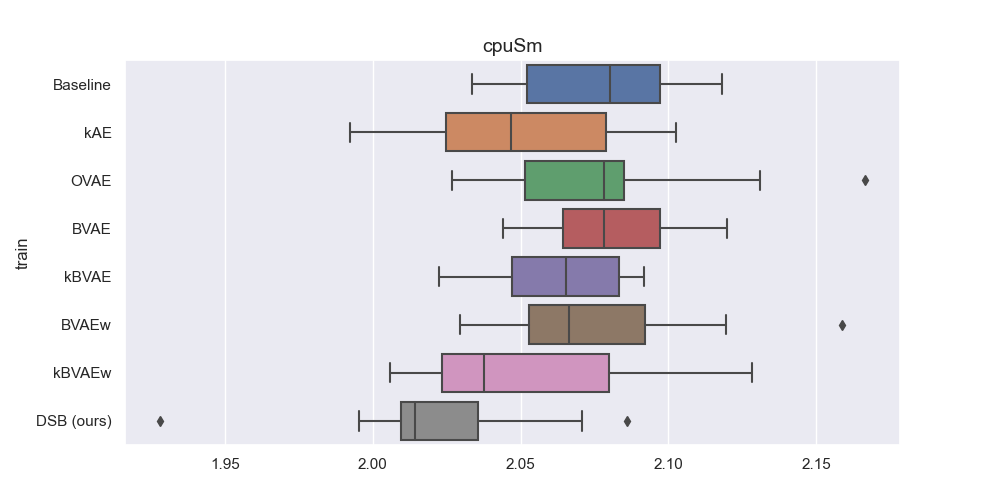}
  % \caption{Our \texttt{KurtHGR} correlation matrix}
  % \label{PearsonMC}
\end{subfigure}
\begin{subfigure}{0.49\textwidth}
  \includegraphics[width=\textwidth]{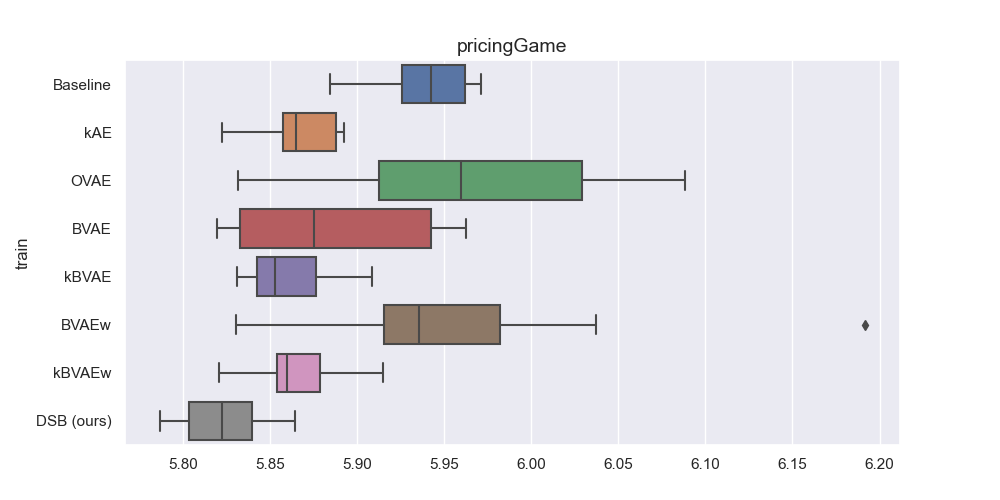}
  % \caption{Our \texttt{KurtHGR} correlation matrix}
  % \label{PearsonMC}
\end{subfigure}
\caption{Ablation study of our approach DSB (in grey)}
\label{Ablation_res}
\end{figure}

As presented in Figure \ref{Benchmark_res} and Table \ref{table_mean}, our Disantengled Deep Smoothed Bootstrap (\textit{DSB}) algorithm outperforms both the initial training sample (baselines) and the current State-of-the-art solutions for all datasets. We therefore applied the non-parametric Wilcoxon test to compare the distributions between our method and the others (k-fold approach). The p-values confirm the significance of the RMSE improvement.  
With ablation study (Figure \ref{Ablation_res}), we observe the interest of our approach.
Lastly, these experiments indicate that using standard synthetic data generators (such as TVAE, CTGAN, and CopGAN) is not recommended for the IR issue. The results here relate to an RMSE, but the conclusions are the same for weighted MSE, R2 or MAE metrics.

% \subsubsection*{Disentanglement weight $\beta_{corr}$ parameter)}
% To analyze the sensitivity of the results to parameter $\beta_{corr}$, we run several models with the following $\beta_{corr}$ values: 0.1, 1, 10, and 100. $\beta_{corr}$ represents the weight of correlation penalization as described in \ref{Final_loss}.  

% Figure \ref{Bcorr_analysis} shows that: 1) The results are quite similar and are better than the baseline  2) The case $\beta_{corr}=1$ presents the best results (green curve of weighted MSE lowest).

% \begin{figure}[H]
% \centering
% \begin{subfigure}{0.49\textwidth}
%          \includegraphics[width=\textwidth]{Imgs/Analyse_Bcorr/wMSE_wY_lmplot.png}
%          \caption{wMSE vs $w_y$ for different $\beta_{corr}$ values}
% \end{subfigure}
% \hfill
% \begin{subfigure}{0.49\textwidth}
%          \includegraphics[width=\textwidth]{Imgs/Analyse_Bcorr/wMSE_wY.png}
%          \caption{Splines of wMSE vs $w_y$ for different $\beta_{corr}$ values}
% \end{subfigure}
% \caption{Prediction weighted MSE vs scarcity ($w_y$) for $\beta_{corr} = 0.1, 1, 10, 100$}
% \label{Bcorr_analysis}
% \end{figure}

\section{Discussion and Perspectives}\label{Discussion}

This paper presents a new way of generating data in the context of imbalanced regression (IR), a challenge that remains relatively unexplored compared to classification, especially for structured data. The key concept is to embed observations into a latent space, enabling more meaningful data generation compared to the original data space. Indeed, generating directly in latent space can be complicated: for example, SMOTE requires that an interpolation between a seed and a nearest neighbor makes sense, which may not be the case, especially in the case of regression, where neighbors are far away. Similarly, applying a smoothed bootstrap or Gaussian perturbation means generating new observations multidimensionally around a seed, without respecting correlations between features, which can introduce bias.
Our empirical results demonstrate that deep learning techniques can be successfully applied to tabular data when appropriately adapted. Indeed, VAEs provide a new representation of data while capturing nonlinear correlations, which is crucial for synthetic data generation. Moreover, this approach enables the handling of mixed (categorical) data, which is challenging to achieve from the original space.
We propose here to leverage the power of VAE inference while rethinking the data generation process. VAEs provide an appropriate framework for using Kernel Density Estimates as they ensure a regular latent space, which may not be the case in the original space, (Kernel-)PCA, or with a vanilla autoencoder. Our \textit{Disentangled Deep Smoothed Bootstrap} is a Smoothed Bootstrap applied on a representative latent space of data. The underlying concept is to utilize the neighborhood of observations in the latent space to more effectively generate rare values, which are inherently difficult to model due to their rarity. The Smoothed bootstrap therefore smoothes the data and avoids the latent space sigmas $\sigma_i$, which are poorly estimated for imbalanced data. 
%Data generation from VAEs is semi-parametric (originating from latent Gaussians that may not always be well-suited). The Deep Smoothed Bootstrap offers non-parametric generation based on a Smoothed Bootstrap, i.e. from a kernel density estimate, requiring less restrictive assumptions than generation from VAE. 
Our simple and effective algorithm yields better results than traditional approaches in IR and than the conventional VAE across multiple datasets with various learning algorithms. Additionally, our study also highlights that using standard synthetic data generators (such as \textit{TVAE} or \textit{CTGAN}) that are not specifically tailored to the problem of IR, is not effective. The main limitation of this approach is that it relies on the quality of the latent representation of the data from the VAE. Indeed, one of the well-known problems of deep learning approaches is the choice of architecture and hyperparameters, and our approach is sensitive to these parameter settings.% \cite{belghazi2018mine,cha2023orthogonality, zhao2019infovae}.
The Deep Smoothed Bootstrap could potentially extend to imbalanced classification, although this area has already been extensively addressed with a plethora of solutions. However, the methodology is more relevant in regression as it allows for handling dependencies between covariates and the target variable.
%It would be particularly interesting to apply this approach to mixed tabular data in the context of Imbalanced Regression, as this remains a relatively uncharted and challenging area, especially due to the correlations associated with qualitative data. 
Lastly, applying this new approach to unstructured data, such as images, within the context of DIR could yield intriguing results. Such experimentation would allow us to assess the effectiveness of the Deep Smoothed Bootstrap in a different domain and potentially uncover novel insights. Obviously, this extension is subject to the condition that the smoothed bootstrap application framework is respected: i) continuous distributions in the latent space and ii) ideally uncorrelated latent distributions.

%Bibliography
\bibliographystyle{unsrt}  
\bibliography{DisDSBCorr}

\end{document}